\documentclass{article}

    \PassOptionsToPackage{numbers, compress}{natbib}
 \usepackage[preprint]{neurips_2026}

\usepackage[utf8]{inputenc} 
\usepackage[T1]{fontenc}    
\usepackage{hyperref}       
\usepackage{url}            
\usepackage{booktabs}       
\usepackage{amsfonts}       
\usepackage{nicefrac}       
\usepackage{microtype}      
\usepackage{xcolor}         
\usepackage{graphicx}
\usepackage{multirow}
\usepackage{amsmath}
\usepackage{amssymb}
\usepackage{enumitem}
\usepackage{algorithm}
\usepackage{algorithmic}
\usepackage{subcaption}
\usepackage{wrapfig}

\newcommand{\mask}{\texttt{[MASK]}}
\newcommand{\dawn}{\texttt{DAWN}}

\title{DAWN: Dependency-Aware Fast Inference for Diffusion LLMs}

\author{%
  Lizhuo Luo$^{1}$ \quad Zhuoran Shi$^{2}$ \quad Jiajun Luo$^{3}$ \quad Zhi Wang$^{3}$\\[1pt]
  \textbf{Shen Ren$^{4}$ \quad Wenya Wang$^{1}$ \quad Tianwei Zhang$^{1}$\thanks{Corresponding author: \texttt{tianwei.zhang@ntu.edu.sg}}}\\[3pt]
  $^{1}$\,Nanyang Technological University \\ $^{2}$\,Southern University of Science and Technology\\
  $^{3}$\,Shenzhen International Graduate School, Tsinghua University \\ $^{4}$\,AUMOVIO Singapore Pte. Ltd.\\
}

\begin{document}

\maketitle

\begin{abstract}
    Diffusion large language models (dLLMs) have shown advantages in text generation, particularly due to their inherent ability for parallel decoding. 
    However, constrained by the quality--speed trade-off, existing inference solutions adopt conservative parallel strategies, leaving substantial efficiency potential underexplored. A core challenge is that parallel decoding assumes each position can be filled independently, but tokens are often semantically coupled. Thus, the correct choice at one position constrains valid choices at others. Without modeling these positional dependencies, parallel strategies produce deteriorated outputs. 
    Motivated by this insight, we propose \textbf{\dawn{}}, a training-free, dependency-aware decoding method for fast dLLM inference. 
    \dawn{} extracts dependency relations and leverages two key observations: (1) masked positions dependent on unmasked certain positions become more reliable; (2) coupled masked positions strongly influence each other's predictions.
    Given those findings, \dawn{} leverages a dependency graph to select more reliable unmasking positions at each iteration, achieving high parallelism with negligible loss in generation quality.
    Extensive experiments across multiple models and datasets demonstrate that \dawn{} speedups the inference by \textbf{1.80 - 8.06$\times$} over baselines while preserving the generation quality. 
    Code is released at \href{https://github.com/lizhuo-luo/DAWN}{https://github.com/lizhuo-luo/DAWN}.
\end{abstract}


\section{Introduction} \label{intro}
Diffusion models have achieved remarkable success in image \citep{sdxl, sana} and video \citep{stablevideo, hunyuanvideo} generation, and have recently been extended to text generation. Unlike autoregressive (AR) models \citep{gpt4, qwen2.5} that generate sequentially, diffusion large language models (dLLMs) \citep{llada, dream} adopt full attention over all positions and refine an entire sequence through multiple denoising iterations, achieving surprisingly strong performance on text generation tasks. dLLMs offer potential solutions to longstanding limitations of AR models, including the inability to decode in parallel and the reversal curse \citep{reversal}. These characteristics have drawn growing research interest in further advancing dLLMs \citep{gemini_diffusion, seeddiffusion, mercury, llada2.0}. 

Despite these efforts, dLLMs still exhibit performance gaps in practical deployments compared to state-of-the-art AR models \citep{fastdllm, eagle3}. These gaps are largely attributed to two main factors: KV-Cache management \citep{fastdllm, dkvcache, dllmcache} and nonindependent position predictions \citep{parallelcurse, fastdllm}. First, dLLMs employ bidirectional attention, which fundamentally contradicts the causal assumption underlying standard KV-Cache mechanisms. Second, the marginal distributions at each position produced by dLLMs often violate the independence assumption underlying parallel decoding.

This work aims to improve the efficiency of parallel decoding in dLLMs, with a particular focus on \textit{nonindependent position predictions}. Most existing parallel decoding strategies \citep{fastdllm, ebsampler} select masked positions using heuristics and relatively coarse-grained criteria, such as confidence and entropy, to ensure that the selected positions behave safely and approximately independently. However, overly conservative selection criteria can substantially limit the achievable parallelism, leaving much of the potential efficiency of parallel decoding underexploited. To address this problem, a natural alternative is to account for positional relationships more directly. Since the main difficulty stems from position coupling, improving parallel decoding requires approximating positional dependencies during inference rather than evaluating each position in isolation. Attention maps \citep{bertlook,dependencyanalysis, sharon, spargeattn, sageattention} provide an approximate yet cheap signal of token interactions that is readily available from each forward pass, making them a practical proxy for dependency estimation. In fact, dependencies between positions of different states yield different effects. From this perspective, two observations are particularly relevant: 
(i) For unmasked-masked position pairs, masked positions that are strongly dependent on previously unmasked high-confidence tokens can remain highly consistent with the final output even when their confidence is relatively low; and (ii) For masked–masked position pairs, committing one position induces a substantial shift in the predictive distribution of its coupled counterpart.
These observations further validate that attention maps can serve as a reasonable proxy for positional interactions and provide a new insight: \textit{dependency-aware inference rules can expand safe parallelism beyond those conservative threshold methods}.

Motivated by this, we propose \dawn{}, a training-free Dependency-AWare fast inference method for diffusioN LLMs. It improves parallel decoding by explicitly accounting for positional dependencies. 
\dawn{} consists of three pipelined components: \textit{Dependency Graph Construction}, \textit{Anchor-Guided Decoding}, and \textit{Conflict-Based Scheduling}. At each denoising iteration, a dependency graph is constructed from the processed attention maps via thresholding to capture salient coupling relations. Based on this graph, the next two components select disjoint sets of positions for parallel updates: $\mathcal{U}_{\text{anchor}}$ and $\mathcal{U}_{\text{conflict}}$. \textit{Anchor-Guided Decoding} first selects high-confidence positions that will probably serve as anchors in subsequent steps, and then previously unmasked high-confidence positions as anchors and relaxes the confidence threshold for strongly anchor-coupled masked positions. Then \textit{Conflict-Based Scheduling} identifies conflicts in the dependency graph and greedily constructs a large non-conflicting update set from the remaining candidates that satisfy a lower confidence threshold. 

Compared with prior approaches that rely on strict positional criteria, \dawn{} relaxes selection constraints by incorporating dependency information, thereby enabling additional updates that would otherwise be filtered out while preserving generation quality. Meanwhile, it substantially mitigates failures caused by nonindependent position predictions. Extensive experiments validate that \dawn{} improves the quality--speed trade-off of dLLM inference across multiple models and settings.

The main contributions are summarized as follows:
\begin{itemize}[leftmargin=*, itemsep=2pt, parsep=2pt, topsep=0pt, partopsep=0pt]


    

    \item We characterize positional dependencies in dLLM inference through attention-derived estimates and identify two key findings: 
    (1) For unmasked--masked position pairs, once a high-confidence token is committed, positions that depend on it become reliably predictable, even at lower confidence; 
    (2) For masked--masked position pairs, committing one position in a coupled pair substantially shifts the predictive distribution of its partner.
    

    \item We propose \textbf{\dawn{}}, a training-free, dependency-aware method for fast inference of diffusion LLMs. It uses the estimated dependencies in two ways: relaxing confidence thresholds for positions anchored by committed high-confidence tokens, and preventing strongly coupled low-confidence positions from being unmasked together, thereby enabling more efficient inference.

    \item Extensive experiments across multiple models, datasets, and representative baselines demonstrate the effectiveness and robustness of \dawn{}, achieving \textbf{1.80 - 8.06$\times$} speedup over the baseline. Ablation studies further validate the contributions of each component and the impact of key hyperparameters. 
\end{itemize}


\section{Preliminaries} \label{prl}
\subsection{Inference Process of dLLMs}
Most recent diffusion LLMs adopt the discrete masked diffusion model paradigm \citep{mdm1, mdm2}, where generation is cast as an iterative unmasking process. Unlike autoregressive models that commit tokens sequentially, dLLMs start from a heavily masked sequence and progressively recover masked positions over several denoising steps until no \mask{} remains. At each step, the model predicts token distributions for all masked positions, conditioned on the current state.

Given a prompt $X$, we initialize the response as a fully masked sequence of predefined length $L$:
\begin{equation*}
y^{(0)} = (\mask{},\ldots,\mask{}) \in (\mathcal{V}\cup\{\mask{}\})^L .
\end{equation*}
Here, $\mathcal{V}$ denotes the vocabulary and $\mask{}$ is the special mask token. In the naive setting, the sampler unmasks exactly one token with the highest confidence per step. At each denoising step $t=0,1,\ldots, L-1$, it concatenates the prompt $X$ and the current response state $y^{(t)}$ as the model input, and commits the corresponding token at the \mask{} response position with the highest confidence:
\begin{equation*}
\begin{aligned}
c_i^{(t)} &\triangleq \max_{v\in\mathcal{V}} p_{\theta}(y_i = v \mid X, y^{(t)}), \quad i\in M^{(t)},\\
i_t &= \arg\max_{i\in M^{(t)}} c_i^{(t)},\\
y_i^{(t+1)} &=
\begin{cases}
\displaystyle \arg\max_{v\in\mathcal{V}} p_{\theta}(y_i = v \mid X, y^{(t)}), & \text{if } i=i_t,\\
y_i^{(t)}, & \text{otherwise.}
\end{cases}
\end{aligned}
\end{equation*}
where $M^{(t)} \triangleq \{ i \mid y^{(t)}_{i}=\mask{} \}$ denotes the set of masked response positions at step $t$. Repeating this procedure for $L$ steps yields a fully unmasked response $y^{(L)}$.

Despite the inherent parallelism within each step, practical dLLM inference exhibits a quality--speed trade-off \citep{d3llm}: directly unmasking multiple tokens in one step leads to substantial quality degradation. 

\subsection{Nonindependent Position Predictions}\label{prl: parallel}
Unmasking a single token per step is inherently slow, and runs counter to the full position computation paradigm of dLLMs, where each denoising step produces predictions for all positions simultaneously. A common approach is to commit tokens at different positions independently according to the model’s predictive distributions at each position:
\begin{equation*}
p_{\theta}\!\left(\{y_i\}_{i\in U^{(t)}} \mid X, y^{(t)}\right) \approx \prod_{i\in U^{(t)}} p_{\theta}\!\left(y_i \mid X, y^{(t)}\right),
\end{equation*}
where $U^{(t)} \subseteq M^{(t)}$ is a set of parallel unmasking positions. However, position predictions in masked refinement are often statistically coupled. Committing multiple tokens at strongly coupled positions in the same step can introduce inconsistencies and degrade generation quality. As the classic example in prior work \citep{parallelcurse, fastdllm}, two-word poker hands (e.g., “high card,” “two pair,” “full house,” “straight flush”) must match as a pair, while parallel sampling can produce invalid combinations such as “high house”. However, if we first decode one of them like “\mask{} house”, then the probability of recovering a valid pair in the next step becomes much higher, which aligns with the conditional probability factorization: $p(y_i, y_j \mid X, y^{(t)}) = p(y_j \mid X, y^{(t)})\, p(y_i \mid y_j, X, y^{(t)})$. This nature of dLLMs suggests that parallel updates should account for position dependencies, avoiding strongly coupled positions while increasing the number of safely updated positions per step.

\subsection{Attention Sinks Bias Dependency Proxies} \label{prl: attnsinks}

Attention sinks \citep{streamllm, attentionsinksdiffusionlanguage} are common in dLLMs. Across multiple methods and diverse samples, we repeatedly observe an abnormal concentration of attention on a small subset of keys. Moreover, the concentration distribution is not static. The specific sink tokens and the strength of aggregation can shift as the denoising step progresses. However, this phenomenon appears largely unrelated to the semantics of sink tokens. We provide a detailed empirical validation of this phenomenon in App.~\ref{apdx: attnsinks}.


Attention sinks bias attention-derived dependency proxies. Those sink positions attract considerable attention and can be misinterpreted as exerting a strong influence on many other tokens. 
\begin{figure*}[t!]
    \centering
    \includegraphics[width=\textwidth]{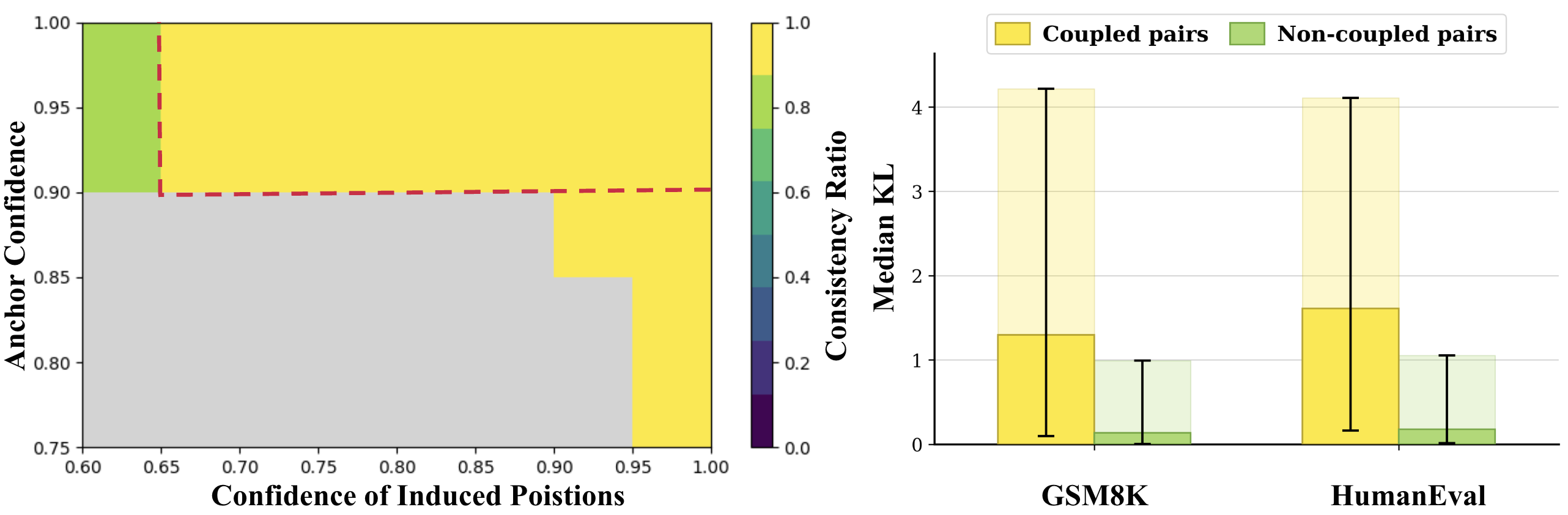}
    \caption{\textbf{Left: Heatmap of Induced Consistency.} Color indicates the consistency ratio (same as the final output) between induced predictions and the final output. Gray cells indicate bins with a negligible fraction of samples and are excluded from analysis. Implementation details are in App.~\ref{apdx: obsdetails1}. \textbf{Right: Influence Between Masked Positions.} Bar height indicates the median KL divergence between a masked position's predictive distribution before and after its paired position is committed, reported separately for coupled and non-coupled pairs. For coupled pairs, the committed position is the source of the directed edge in the dependency graph; for non-coupled pairs, it is chosen at random. Error bars denote the [25\%--75\%] percentile range. Implementation details are in App.~\ref{apdx: obsdetails2}. Additional results are in App.~\ref{apdx:additional_obs}, further validating this observation.}
    \label{fig:obs}
\end{figure*}

\section{Observations} \label{observation}

The effect of positional dependencies varies with the position state. We examine two cases: dependencies between unmasked--masked position pairs, and between masked--masked position pairs. Details follow in the subsequent subsections.

\subsection{High-Confidence Unmasked Positions as Anchors} \label{obs: anchors}

For unmasked-masked position pairs. During dLLM inference, masked positions can exhibit high consistency even when their confidence is not particularly high. Inspired by LocalLeap \citep{localleap}, we find that the induced positions that are strongly dependent on prompts or previously unmasked tokens with high confidence (anchors) often exhibit high consistency despite relatively low instantaneous confidence. This is validated by the highlighted region in the upper area of the left of Fig.~\ref{fig:obs}: the induced positions can remain consistent with the final output when their confidence is low. In particular, when the corresponding anchors have confidence above 0.9, the induced tokens are consistently correct at a relatively low confidence. Therefore, the safety of parallel updates depends not only on confidence but also on whether the prediction is sufficiently conditioned on reliable context. 

\subsection{Coupled Masked Positions Exhibit Strong Influence} \label{obs: cbs}

For masked--masked position pairs, since we represent positional relationships as a directed graph derived from attention maps (details at Sec.~\ref{meth: dgc}), the right of Fig.~\ref{fig:obs} shows that committing the source of a directed edge in a coupled pair induces a much larger KL shift in the target's predictive distribution than committing a randomly chosen position in a non-coupled pair. This suggests that sequentially decoding coupled positions yields more consistent joint outputs (the remaining position will be influenced), alleviating the \textit{nonindependent parallel predictions} issue. It also validates the use of attention as a dependency proxy.
\section{Methodology}\label{method}

\begin{figure*}[t]
    \centering
    \includegraphics[width=\textwidth]{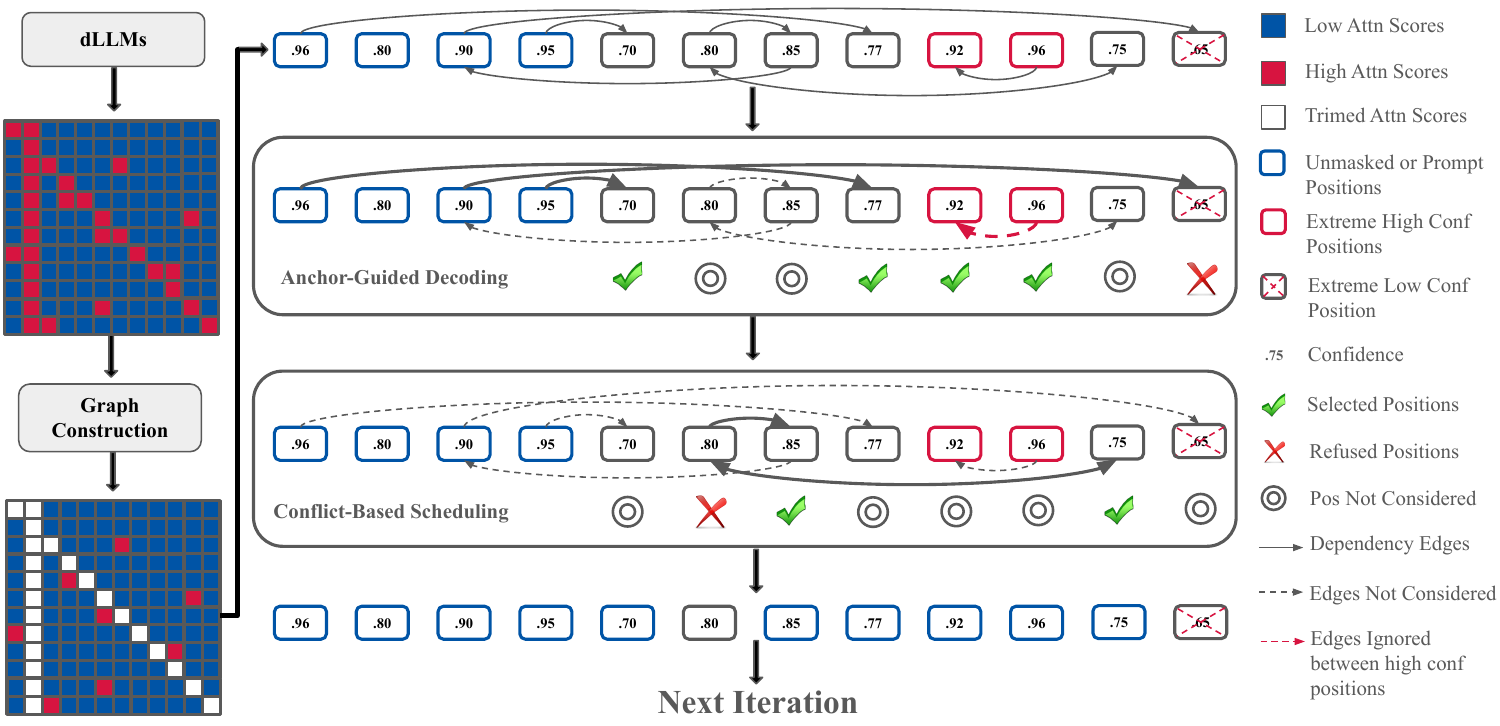}
    \caption{\textbf{Overview of \dawn{}.}
    Left: \textit{Dependency Graph Construction} preprocesses the attention map and extracts a sparse directed dependency graph by retaining only salient (high-score) attention links.
    Middle: guided by this graph, \textit{Anchor-Guided Decoding} and \textit{Conflict-Based Scheduling} select two sets of positions, and the union of selected positions is unmasked simultaneously.}
    \label{fig:arch}
    \vspace{-10pt}
\end{figure*}

Driven by the observations, we propose \dawn{}, a training-free, dependency-aware solution to accelerate dLLM inference while maintaining the generation quality.  
The key idea is to \textit{extract positional dependencies and leverage them to select a larger set of reliable positions for update at each iteration}.
\dawn{} realizes efficient inference by three pipelined modules, as shown in Fig.~\ref{fig:arch}.
\textit{Dependency Graph Construction} builds a sparse directed graph from a lightweight proxy of token dependencies based on attention maps \citep{bertlook, dependencyanalysis, sharon}. \textit{Anchor-Guided Decoding} first selects high-confidence masked positions that can serve as anchors for the next forward and are approximately reliable \citep{fastdllm}, and then leverages high-confidence prompts or committed positions as anchors, enabling strongly coupled positions (induced positions) to be unmasked with a relatively low confidence. These together produce a parallel set $\mathcal{U}^{(t)}_{\text{anchor}}$. For the remaining tokens, \textit{Conflict-Based Scheduling} uses conflict relations induced by the dependency graph to select a maximum independent set $\mathcal{U}^{(t)}_{\text{conflict}}$ from positions that meet a lower confidence threshold. Finally, the positions in  $\mathcal{U}^{(t)}_{\text{anchor}}\cup~\mathcal{U}^{(t)}_{\text{conflict}}$ are unmasked in parallel in this iteration. These three modules are pipelined within each denoising iteration to enable accurate and efficient inference. The following subsections describe their mechanisms in detail.

\subsection{Dependency Graph Construction}\label{meth: dgc}
Prior studies \citep{bertlook, dependencyanalysis, sharon} indicate that attention maps capture information flow and dependency relations among tokens. Meanwhile, many existing works \citep{fastv, spargeattn, sageattention, svg} use attention maps to characterize token interactions, thereby enabling more efficient inference. Inspired by these, we treat attention maps as a lightweight, approximate proxy for token dependencies during decoding. This coupling signal is readily available from each forward pass and can be converted into a sparse directed dependency graph, which serves as the basis for subsequent efficient scheduling.

Since attention patterns evolve across denoising iterations, the dependency graph is constructed at each iteration from the model's attention maps. \citet{dependencyanalysis} shows that deeper layers tend to encode more semantics. So to obtain a signal that is closer to the semantics and final prediction, attention weights are averaged across the last few layers and all heads. As discussed in Section~\ref{prl: attnsinks}, dLLMs often exhibit attention sinks, where a small set of positions absorbs most of the attention mass due to systematic bias rather than semantic relevance. This can lead to misleading dependent relations when attention is used as a proxy for dependency. To mitigate this effect, sink positions are identified via outlier detection and filtering of their incoming attention mass: given an aggregated attention matrix $A^{(t)}\in\mathbb{R}^{L\times L}$ at iteration $t$, the incoming attention mass of position $j$ is defined as
\begin{equation*}
\bar{A}^{(t)}_j = \frac{1}{L}\sum_{i=1}^{L} A^{(t)}_{i,j},
\end{equation*}
and position $j$ is marked as a sink if $\bar{A}^{(t)}_j$ is larger than a predefined threshold $\tau_{sink}$. Self-attention on the diagonal is ignored as it does not capture cross-position dependencies.

Given the processed attention proxy at iteration $t$, a directed sparse dependency graph is constructed to capture salient token couplings during decoding. 
Specifically, to keep the graph sparse and focus on the most informative relations, edges are retained based on thresholded ($\tau_{edge}$) attention scores. A directed edge $j\!\rightarrow\! i$ is added if query position $i$ places a sufficiently large attention mass on key position $j$, indicating that the prediction at token $i$ can be significantly conditioned by token $j$. 
 The resulting graph provides an approximate representation of positional dependencies and will be used by subsequent modules.

\subsection{Anchor-Guided Decoding}
In fact, some low-confidence tokens at masked positions can still be consistent with the final response, indicating that focusing only on high-confidence positions remains a limitation for dLLM inference. Motivated by this, \textit{Anchor-Guided Decoding} is introduced to accept approximately independent positions and reliable positions under a relaxed confidence criterion. 

As discussed in Sec.~\ref{obs: anchors}, low-confidence positions that are strongly coupled to previously unmasked positions meet the confidence threshold 0.9 and often match the final result. Following this observation, we define anchors as positions that are prompts or have been unmasked and whose confidence meets $\tau_{high}$, where $\tau_{high}$ is observed as 0.9. For each step, we first select positions that meet $\tau_{high}$ and will likely serve as anchors in subsequent steps, which is also supported by prior work~\citep{fastdllm} showing that parallel decoding remains relatively stable among sufficiently high-confidence positions. Then, using the dependency graph, we identify the induced positions $\mathcal{I}$ as masked positions that are reachable via directed edges from anchor tokens. Intuitively, higher-confidence anchors provide more reliable context, allowing the corresponding induced positions to be unmasked at lower confidence. Accordingly, the set of positions eligible to be unmasked at the denoising step $t$ is defined as:
\begin{equation*}
\mathcal{U}^{(t)}_{\text{anchor}}
=
\left\{
\, i \;\middle|\;
\begin{array}{l}
i \in M^{(t)},\ c_i \ge \tau_{\text{high}}
\ \text{or}\\
i \in \mathcal{I}^{(t)},\ c_i \ge \tau_{\text{induced}}
\end{array}
\right\}.
\end{equation*}
where $\tau_{induced}$ is the confidence threshold for unmasking induced positions. 
Overall, \textit{Anchor-Guided Decoding} selects approximately reliable positions under a high confidence threshold and relaxes the threshold for induced positions, enabling more efficient inference.

\subsection{Conflict-Based Scheduling}

\begin{wrapfigure}{r}{0.5\columnwidth}
\vspace{-\intextsep}      
\begin{minipage}{\linewidth}
\begin{algorithm}[H]      
  \caption{Conflict-Based Scheduling}
  \label{alg:conflict}
  \small                  
  \begin{algorithmic}[1]
    \STATE {\bfseries Input:} remaining candidate positions $\mathcal{C}^{(t)}$, anchor update set $\mathcal{U}^{(t)}_{\text{anchor}}$, confidence scores $\{c_i\}_{i\in\mathcal{C}^{(t)}}$, conflict neighbors $\{\mathcal{N}(i)\}_{i\in\mathcal{C}^{(t)}}$, lower threshold $\tau_{\mathrm{low}}$
    \STATE {\bfseries Output:} parallel update set $\mathcal{U}^{(t)}_{\text{conflict}}$
    \STATE Initialize $\mathcal{U}^{(t)}_{\text{conflict}} \leftarrow \emptyset$
    \STATE $\mathcal{X} \leftarrow \mathcal{U}^{(t)}_{anchor} \cup \bigcup_{i \in \mathcal{U}^{(t)}_{anchor}} \mathcal{N}(i)$ \COMMENT{selected positions and their conflicts}
    \STATE $\mathcal{R} \leftarrow \{\, i \in \mathcal{C}^{(t)} \mid c_i \ge \tau_{\mathrm{low}} \,\} \setminus \mathcal{X}$ \COMMENT{remaining candidates}
    \WHILE{$\mathcal{R} \neq \emptyset$}
      \STATE Select $i^\star \leftarrow \arg\max_{i \in \mathcal{R}} c_i$
      \STATE $\mathcal{U}^{(t)}_{\text{conflict}} \leftarrow \mathcal{U}^{(t)}_{\text{conflict}} \cup \{i^\star\}$
      \STATE $\mathcal{R} \leftarrow \mathcal{R} \setminus \left(\{i^\star\} \cup \mathcal{N}(i^\star)\right)$
    \ENDWHILE
    \STATE {\bfseries return} $\mathcal{U}^{(t)}_{\text{conflict}}$
  \end{algorithmic}
\end{algorithm}
\end{minipage}
\end{wrapfigure}

For the remaining low-confidence positions that are strongly coupled and satisfy a lower confidence threshold $\tau_{low}$, simultaneously unmasking them may introduce inconsistent commitments, as discussed in Sec.~\ref{prl: parallel}. To mitigate such errors while retaining high parallelism, \textit{Conflict-Based Scheduling} is introduced to prevent low-confidence but highly coupled positions from being decoded in the same step.

Since the dependency graph captures salient positional dependencies, it can be used to identify strongly coupled position pairs. Sec.~\ref{obs: cbs} indicates that strongly coupled masked positions exert significant influence. By first decoding the source position in a coupled pair, the other can be predicted conditioned on the committed position at the next step, yielding more consistent outputs and mitigating the inconsistency introduced by nonindependent parallel predictions if update them together. Therefore, a conflict relation is defined between two positions connected by an edge in the dependency graph, regardless of direction: if either $i\rightarrow j$ or $j\rightarrow i$ is present, then positions $i$ and $j$ are considered to be in conflict and should not be unmasked simultaneously.

Algorithm~\ref{alg:conflict} shows the detailed procedure. Specifically, at iteration $t$, based on these conflicts and the graph topology, a greedy independent set is constructed from the remaining positions that (i) are not yet included in and not conflicting neighbors of ~$\mathcal{U}^{(t)}_{\text{anchor}}$, and (ii) satisfy $\tau_{low}$, yielding additional positions that can be decoded in parallel. Concretely, positions are greedily selected in descending order of confidence: the highest-confidence position is added to the update set $\mathcal{U}^{(t)}_{\text{conflict}}$, and all of its conflicting neighbors are removed from further consideration. This procedure repeats until no candidate positions remain.


In practice, quality degradation from naively lowering the confidence threshold is largely driven by positional coupling. By explicitly avoiding simultaneous unmasking of strongly coupled positions, \textit{Conflict-Based Scheduling} helps maintain decoding quality while allowing a lower confidence threshold $\tau_{low}$, thus speeding up the inference.

\section{Experiments} \label{exp}

\begin{table*}[t!]
  \centering
  \small
  \caption{\textbf{Performance comparison between \dawn{} and baselines across 4 datasets and 4 models}. We report Accuracy, TPS, and Speedup to assess their efficiency and generation quality.}
  \label{tab:main_results}
  \resizebox{\textwidth}{!}{
    \begin{tabular}{c*{4}{ccc}}
      \toprule
        & \multicolumn{3}{c}{GSM8K}
        & \multicolumn{3}{c}{MATH}
        & \multicolumn{3}{c}{HumanEval}
        & \multicolumn{3}{c}{MBPP}\\
      \cmidrule(lr){2-4} \cmidrule(lr){5-7} \cmidrule(lr){8-10} \cmidrule(lr){11-13} 
      Method
        & Acc.$\uparrow$ & TPS$\uparrow$ & Speedup$\uparrow$
        & Acc.$\uparrow$ & TPS$\uparrow$ & Speedup$\uparrow$
        & Acc.$\uparrow$ & TPS$\uparrow$ & Speedup$\uparrow$
        & Acc.$\uparrow$ & TPS$\uparrow$ & Speedup$\uparrow$\\
      \midrule
      \multicolumn{13}{c}{LLaDA-8B-Instruct}\\
      \midrule
      Vanilla & 77.94 & 10.32 & 1.00$\times$ & 33.20 & 14.10 & 1.00$\times$ & 40.24 & 26.46 & 1.00$\times$ & 29.60 & 9.46 & 1.00$\times$\\
      Confidence & 78.39 & 33.44 & 3.24$\times$ & 32.98 & 36.88 & 2.62$\times$ & 40.85 & 86.73 & 3.28$\times$ & 30.00 & 34.81 & 3.68$\times$\\
      KLASS & 76.12 & 23.23 & 2.25$\times$ & 32.32 & 25.27 & 1.79$\times$ & 36.59 & 56.91 & 2.15$\times$ & 27.00 & 22.62 & 2.39$\times$\\
      LocalLeap & 77.33 & 44.19 & 4.28$\times$ & 32.42 & 47.12 & 3.34$\times$ & 39.63 & \textbf{109.80} & \textbf{4.15$\times$} & 30.80 & 44.13 & 4.66$\times$\\
      \dawn{} & 77.94 & \textbf{44.72} & \textbf{4.33$\times$} & 32.36 & \textbf{48.16} & \textbf{3.42$\times$} & 40.24 & 108.99 & 4.12$\times$ & 30.80 & \textbf{45.17} & \textbf{4.77$\times$} \\
      \midrule
      \multicolumn{13}{c}{LLaDA-1.5}\\
      \midrule
      Vanilla & 81.12 & 9.65 & 1.00$\times$ & 33.36 & 12.75 & 1.00$\times$ & 43.90 & 9.26 & 1.00$\times$ & 38.80 & 3.45 & 1.00$\times$\\
      Confidence & 80.74 & 32.49 & 3.37$\times$ & 33.38 & 32.87 & 2.58$\times$ & 42.68 & 23.39 & 2.53$\times$ & 38.80 & 20.40 & 5.91$\times$\\
      KLASS & 77.63 & 22.63 & 2.35$\times$ & 31.80 & 23.29 & 1.83$\times$ & 39.63 & 16.49 & 1.78$\times$ & 30.00 & 13.25 & 3.84$\times$\\
      LocalLeap & 80.14 & 42.60 & 4.41$\times$ & 32.44 & 42.36 & 3.32$\times$ & 42.07 & \textbf{30.21} & \textbf{3.26$\times$} & 39.20 & 27.00 & 7.83$\times$\\
      \dawn{} & 80.82 & \textbf{43.24} & \textbf{4.48$\times$} & 31.80 & \textbf{42.84} & \textbf{3.36$\times$} & 42.07 & 29.53 & 3.19$\times$ & 37.60 & \textbf{27.80} & \textbf{8.06$\times$} \\
      \midrule
      \multicolumn{13}{c}{Dream-v0-Base-7B}\\
      \midrule
      Vanilla & 76.42 & 14.37 & 1.00$\times$ & 34.70 & 18.16 & 1.00$\times$ & 38.41 & 33.66 & 1.00$\times$ & 53.40 & 11.85 & 1.00$\times$\\
      Confidence & 73.84 & 23.41 & 1.63$\times$ & 34.38 & 44.66 & 2.46$\times$ & 39.63 & 61.54 & 1.83$\times$ & 53.80 & 50.37 & 4.25$\times$\\
      KLASS & 72.00 & 12.58 & 0.88$\times$ & 32.80 & 23.16 & 1.27$\times$ & 42.68 & 31.32 & 0.93$\times$ & 55.80 & 19.46 & 1.64$\times$\\
      LocalLeap & 72.55 & 25.40 & 1.77$\times$ & 34.48 & 51.32 & 2.83$\times$ & 36.59 & 68.28 & 2.03$\times$ & 53.60 & 59.38 & 5.01$\times$\\
      \dawn{} & 73.54 & \textbf{25.88} & \textbf{1.80$\times$} & 34.00 & \textbf{51.45} & \textbf{2.83$\times$} & 39.63 & \textbf{68.96} & \textbf{2.05$\times$} & 53.20 & \textbf{64.55} & \textbf{5.45$\times$} \\
      \midrule 
      \multicolumn{13}{c}{Dream-v0-Instruct-7B}\\
      \midrule
      Vanilla & 76.35 & 7.30 & 1.00$\times$ & 37.80 & 17.73 & 1.00$\times$ & 53.66 & 22.66 & 1.00$\times$ & 54.20 & 8.37 & 1.00$\times$\\
      Confidence & 75.20 & 28.81 & 3.95$\times$ & 38.24 & 37.92 & 2.14$\times$ & 57.32 & 52.11 & 2.30$\times$ & 55.60 & 39.09 & 4.67$\times$\\
      KLASS & 71.72 & 12.93 & 1.77$\times$ & 34.66 & 18.74 & 1.06$\times$ & 54.88 & 26.71 & 1.18$\times$ & 56.80 & 18.53 & 2.21$\times$\\
      LocalLeap & 73.24 & 32.94 & 4.51$\times$ & 38.10 & 43.66 & 2.46$\times$ & 54.27 & 59.28 & 2.62$\times$ & 55.00 & 42.71 & 5.10$\times$\\
      \dawn{} & 73.16 & \textbf{32.99} & \textbf{4.52$\times$} & 38.22 & \textbf{44.18} & \textbf{2.49$\times$} & 54.88 & \textbf{60.23} & \textbf{2.66$\times$} & 55.80 & \textbf{44.03} & \textbf{5.26$\times$} \\
      \bottomrule
    \end{tabular}
  }
\end{table*}

\subsection{Setups}
\noindent \textbf{Models and Benchmarks.} We evaluate our approach on several variants of two models: LLaDA-8B-Instruct \citep{llada}, LLaDA-1.5 \citep{llada1.5}, Dream-v0-Base-7B \citep{dream}, Dream-v0-Instruct-7B. Benchmarks include diverse datasets: GSM8K (5-shot) \citep{gsm8k}, MATH (4-shot) \citep{math}, HumanEval (0-shot) \citep{humaneval}, and MBPP (3-shot) \citep{mbpp}, covering a range of reasoning and code generation tasks. 
We report tokens per second (TPS), relative speedup ratio (Speedup) and number of function evaluations (NFE) to reflect efficiency, along with task accuracy (Acc.).

\noindent \textbf{Baselines.} We compare \dawn{} against four baselines: the Vanilla sampling method (Top-1 Sampling), which unmasks the top-1 confidence position at each iteration; the Confidence-Aware Parallel proposed by Fast-dLLM \citep{fastdllm}, selecting positions whose confidence exceeds a predefined threshold; KLASS \citep{klass}, using both confidence and KL divergence to select positions; and LocalLeap \citep{localleap}, identifying anchors and performing localized relaxed parallel decoding. 

\noindent \textbf{Hardware and Implementation Details.} Our experiments are conducted on an NVIDIA H100 80G GPU. We set the generation length to 256 and the block length to 32 for all methods except KLASS, which uses its best-performing block length. All baselines are evaluated under their default hyperparameter settings. For \dawn{}, $\tau_{high}$ is set to 0.9 according to our observation and Fast-dLLM. $\tau_{low}$ is set from 0.7 to 0.85. Full configurations and the justifications can be found in App.~\ref{apdx: hyper}. All evaluations are conducted using the standardized lm-eval \citep{lm-eval} library. KV cache integration is trivial and would yield further speedup, but we omit it for consistent comparison with other works.

\subsection{Main Results}

\begin{figure}[t!]
    \centering
    \includegraphics[width=\columnwidth]{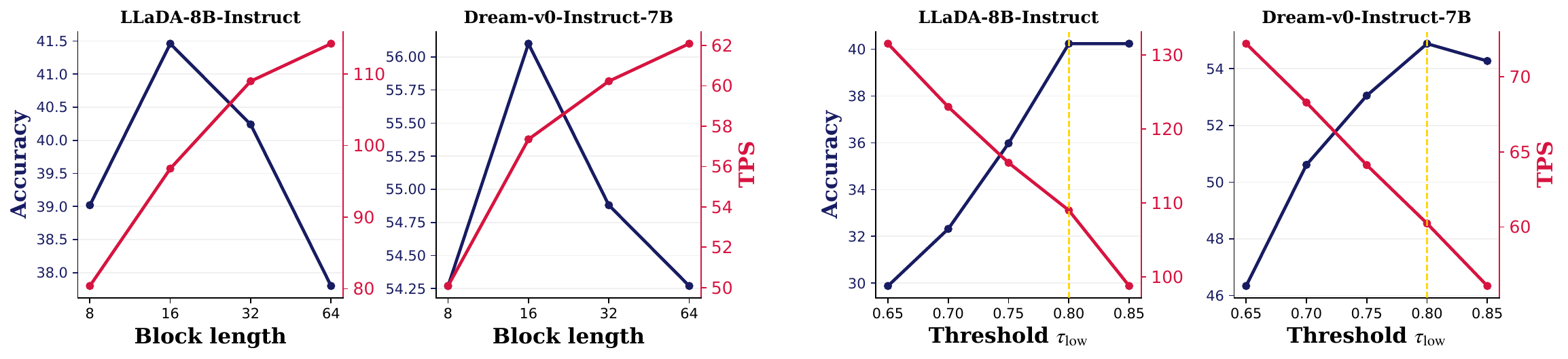}
    \caption{\textbf{Left: Effectiveness of \dawn{} on HumanEval under different block lengths}. We report accuracy (blue, left y-axis) and TPS (red, right y-axis). \textbf{Right: We vary the lower threshold $\tau_{low}$ and report accuracy (blue, left y-axis) and throughput TPS (red, right y-axis) on HumanEval}. The dashed line (yellow) marks the default setting $\tau_{low}=0.80$.}
    \label{fig:blen_and_low}
\end{figure}

Tab.~\ref{tab:main_results} reports the accuracy and efficiency of each method across different models on four benchmarks.

Overall, \dawn{} achieves a substantial improvement in inference speed while maintaining accuracy comparable to or even slightly higher than the vanilla method. \dawn{} achieves substantially higher throughput than prior decoding baselines, with up to 8.06$\times$ speedup on MBPP using the LLaDA-1.5 model. Moreover, these gains do not come at the cost of quality. On LLaDA-8B-Instruct, \dawn{} achieves 77.94 accuracy on GSM8K, matching the vanilla method, and attains 30.80 on MBPP, slightly higher than the vanilla 29.60. In other settings, \dawn{} incurs only negligible quality loss, indicating a more favorable quality--speed trade-off. 

Compared with confidence-aware parallel and KLASS, \dawn{} significantly improves throughput while achieving nearly identical accuracy. Compared with LocalLeap, \dawn{} shows advantages in some cases: it improves throughput by approximately 0.05 – 5.17 tokens per second while enhancing accuracy by up to 3.04. While the overall performance of the two methods is often comparable, they differ fundamentally in both motivation and methodology. LocalLeap leverages local determinism around high-confidence tokens to accelerate decoding, whereas \dawn{} explicitly models positional dependencies through attention-derived graphs and exploits them for both \textit{Anchor-Guided Decoding} and \textit{Conflict-Based Scheduling}. Further evaluation and statistical results are in App.~\ref{apdx:ifeval} and App.~\ref{statistical}.



\subsection{Ablation Study}


\begin{wraptable}{r}{0.5\columnwidth}
\vspace{-4pt}
  \centering
  \small
  \caption{\textbf{Ablation study on the effectiveness of key modules in \dawn{} across 2 datasets and 2 models}. We report Accuracy, TPS, and NFE to assess their quality and speed. The settings are the same as in the main experiments.}\label{tab:ablation}
  \resizebox{\linewidth}{!}{
    \begin{tabular}{c*{2}{ccc}}
      \toprule
        & \multicolumn{3}{c}{GSM8K}
        & \multicolumn{3}{c}{HumanEval}\\
      \cmidrule(lr){2-4} \cmidrule(lr){5-7}
      Method
        & Acc.$\uparrow$ & TPS$\uparrow$ & NFE$\downarrow$ 
        & Acc.$\uparrow$ & TPS$\uparrow$ & NFE$\downarrow$  \\
      \midrule
      \multicolumn{7}{c}{LLaDA-8B-Instruct}\\
      \midrule
      \dawn{} & 77.94 & \textbf{44.72} & \textbf{55.76} & 40.24 & \textbf{109.0} & \textbf{61.63} \\
      + CBS & 76.80 & 22.31 & 112.9 & \textbf{41.46} & 51.51 & 131.0 \\
      + AGD & \textbf{78.47} & 33.83 & 74.69 & 41.46 & 92.52 & 72.93 \\
      Vanilla & 77.94 & 10.32 & 256 & 40.24 & 26.46 & 256 \\
      \midrule 
      \multicolumn{7}{c}{Dream-v0-Instruct-7B}\\
      \midrule
      \dawn{} & 73.16 & \textbf{32.99} & \textbf{51.92} & 54.88 & \textbf{60.23} & \textbf{77.12} \\
      + CBS & 73.31 & 29.33 & 58.13 & 54.88 & 54.79 & 83.34 \\
      + AGD & 75.28 & 27.63 & 62.19 & \textbf{57.31} & 51.97 & 93.01 \\
      Vanilla & \textbf{76.35} & 7.30 & 256 & 53.66 & 22.66 & 256 \\
      \bottomrule
    \end{tabular}
  }
\end{wraptable}
We conduct extensive ablation studies to assess the contributions of key components in \dawn{}. We further examine the sensitivity of \dawn{} to the block length, the lower confidence threshold $\tau_{low}$, and the generation length to understand how these factors affect its effectiveness. 
All other settings follow the main experiment.

\noindent \textbf{Effectiveness of key components.} Since \textit{Dependency Graph Construction} serves as the foundation for the other two modules, we focus on validating the effectiveness of components for selecting update set. Tab.~\ref{tab:ablation} summarizes ablations of \dawn{} on two representative dLLMs and benchmarks. In particular, adding \textit{Anchor-Guided Decoding} (AGD) alone on top of the vanilla baseline substantially improves throughput while maintaining or even slightly improving accuracy. On LLaDA-8B-Instruct, TPS rises from $10.32$ to $33.83$ on GSM8K while accuracy improves from $77.94$ to $78.47$, indicating that AGD accelerates inference with virtually no accuracy loss. Adding \textit{Conflict-Based Scheduling} (CBS) alone instead yields strong efficiency gains at the cost of a small accuracy drop. On Dream-v0-Instruct-7B, CBS even outperforms AGD in speedup, lifting TPS from $22.66$ to $54.79$ on HumanEval. Overall, the full solution consistently achieves the best efficiency while maintaining comparable accuracy, demonstrating that both modules contribute significantly to the inference acceleration.



\noindent \textbf{Impact of block length.} The left of Fig.~\ref{fig:blen_and_low} reports the accuracy and throughput of \dawn{} with different block lengths. As the block length increases, throughput increases for both methods as higher parallelism, but accuracy first increases and then decreases. Overall, it preserves a robust quality-speed trade-off across a wide range of lengths.


\begin{wrapfigure}{r}{0.5\columnwidth}
    \centering
    \vspace{-5pt}
    \includegraphics[width=0.5\columnwidth]{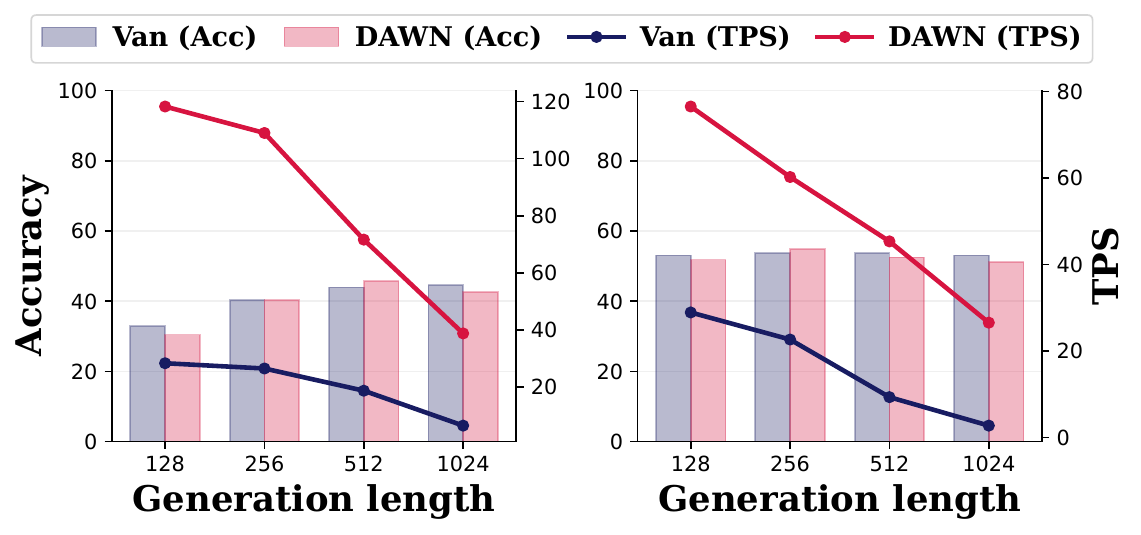}
    \caption{\textbf{Effectiveness of \dawn{} and the vanilla on HumanEval under different generation lengths}. Bars report accuracy (left y-axis) and solid lines report TPS (right y-axis).}
    \label{fig:len_abl}
    \vspace{-13pt}
\end{wrapfigure}

\noindent \textbf{Impact of lower threshold.} The right of Fig.~\ref{fig:blen_and_low} reports the evaluation under different values of lower threshold $\tau_{low}$. We observe a clear quality-speed trade-off on both models. Reducing $\tau_{low}$ increases TPS by admitting more parallel updates, but can hurt accuracy due to less reliable low-confidence commitments. 
The default $\tau_{low}=0.80$ (dashed line) maintains a high generation quality and improved efficiency.

\noindent \textbf{Impact of generation length.} Fig.~\ref{fig:len_abl} reports the accuracy and TPS of \dawn{} and the vanilla with different generation lengths $L$. Across both LLaDA and Dream models, \dawn{} consistently improves TPS over the vanilla while maintaining comparable accuracy. As $L$ increases, TPS decreases for both methods due to higher denoising costs, yet \dawn{} continues to boost the efficiency. Overall, it preserves a more favorable quality-speed trade-off across a wide range of lengths.

\section{Related Work} \label{related work}

\noindent \textbf{Diffusion Large Language Models.}
Recently, diffusion language models have gained renewed interest and emerged as a competitive alternative for text generation, demonstrating promising progress across various tasks.
Representative approaches include pretraining dLLMs from scratch \citep{llada} and building dLLMs on top of existing AR models \citep{dream}. In parallel, several commercial systems \citep{seeddiffusion, mercury, gemini_diffusion} highlight the feasibility and practical potential of diffusion-based generation. Recent works \citep{fastdllmv2, llada2.0, wedlm} continue to advance dLLMs along multiple dimensions, including scaling to larger model size and exploring alternative training-inference paradigms for faster refinement. Moreover, diffusion-based modeling has been extended to multimodal settings, where multimodal dLLMs \citep{llada-v, mmada} demonstrate strong performance across various tasks and point toward more unified diffusion-based generative models.

\noindent \textbf{Efficient Inference of dLLMs.}
Despite dLLMs' ability to parallel decode, practical challenges remain during inference, motivating more efforts on faster and more reliable decoding. 
Existing works focus on the KV Cache optimization \citep{fastdllm, dkvcache}, early stopping \citep{dllm-var, daedal}, distillation-based \citep{dparallel} acceleration, and others \citep{quantdllm, sparsedllm, dsb}. 
These directions have shown promising gains in improving the efficiency of dLLM inference. Beyond these optimizations, numerous studies have explored optimized sampling strategies for dLLM inference. Fast-dLLM \citep{fastdllm} adopts a confidence-aware strategy that unmasks multiple positions when their scores are sufficiently high, thereby making the independence approximation more reliable. EB-Sampler \citep{ebsampler} uses the entropy of predictive distributions to decide which positions are safe to update in parallel. KLASS \citep{klass} further incorporates temporal stability by comparing distributions across iterations via the KL divergence. 
WINO \citep{wino} follows a draft-and-verify style: it drafts many tokens in parallel and selectively regenerates those that fail verification. 
Spiffy and related works \citep{spiffy, orchestrating} apply speculative decoding style strategies to dLLMs to accelerate diffusion decoding. LocalLeap \citep{localleap} leverages a local determinism hypothesis, observing that positions adjacent to high-confidence commits tend to stabilize earlier and can therefore be updated more aggressively. 
Unlike the above methods, this work explicitly approximates token coupling during inference and leverages this approximation to guide efficient parallel sampling.
\section{Limitations} \label{limitation}

\dawn{} introduces several hyperparameters that require tuning to achieve a favorable performance. However, as discussed in App.~\ref{apdx: hyper}, performance is primarily sensitive to $\tau_{low}$, while the remaining parameters can adopt fixed defaults for each model. This reduces the practical tuning burden.



\section{Conclusion} \label{conclusion}
This work leads to a better quality-speed trade-off for dLLM inference, narrowing the gap to state-of-the-art language models in practical generation settings. Specifically, we propose \textbf{\dawn{}}, a training-free, dependency-aware fast inference method for dLLMs.
It is primarily motivated by the inherent challenge of nonindependent position predictions in dLLMs and mitigates this issue by adopting a positional-dependency perspective, offering a complementary approach to alleviate failures in parallel unmasking.
Guided by a sparse directed dependency graph, \dawn{} selects unmasking positions at each iteration and enables highly parallel updates while preserving generation quality. 
Extensive experiments across multiple models and datasets validate the effectiveness of \dawn{}, demonstrating consistent speedups while maintaining comparable quality.

\begin{ack}
    This research/project is supported by A*STAR under the RIE2025 Industry Alignment Fund – Industry Collaboration Projects (IAF-ICP) Funding Initiative (Award: I2501E0045 (H25-MCP0826)), as well as cash and in-kind contribution from the industry partner(s). 
    
    The authors declare no competing interests.
\end{ack}



\bibliographystyle{plainnat}
\small
\bibliography{main}


\appendix
\onecolumn
\section{Attention Sinks Validation} \label{apdx: attnsinks}

\begin{figure*}[h]
    \centering
    \includegraphics[width=\textwidth]{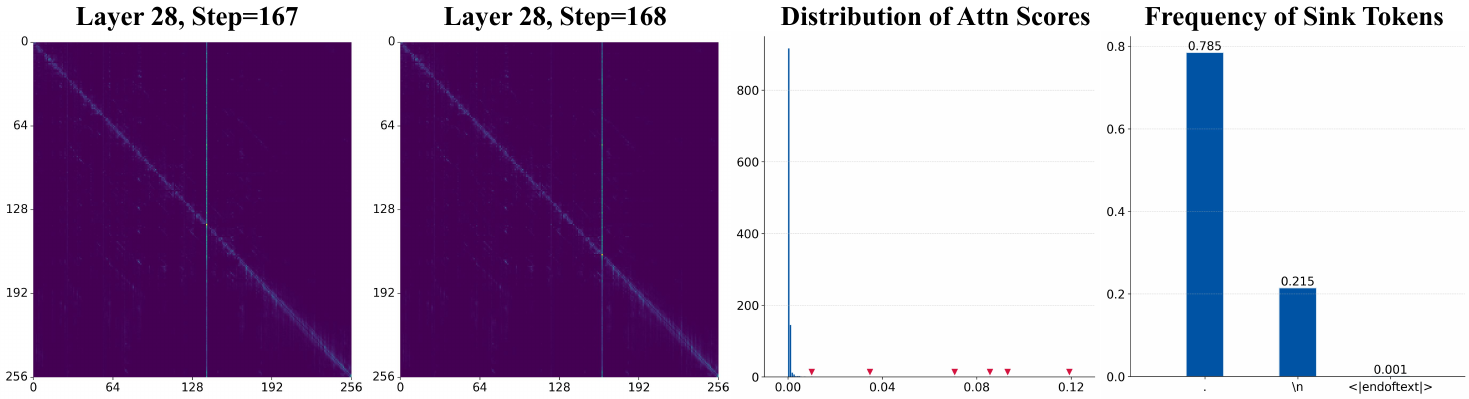}
    \caption{\textbf{Attention Sinks in dLLMs.} We conduct experiments on multiple samples using LLaDA-8B-Instruct. Left: The two heatmaps show partial attention maps from the same layer at different denoising steps, illustrating that the attention sink shifts across denoising iterations. Right: The third plot reports the distribution of attention scores corresponding to the first plot, and the rightmost plot reports the frequency of sink tokens from multiple samples.}
    \label{fig:attn_sink}
\end{figure*}

As shown in Fig.~\ref{fig:attn_sink}, each heatmap exhibits an abnormal concentration of attention mass. Comparing attention maps across denoising steps further indicates that the sink location shifts over iterations. In the rightmost plot, most identified sink tokens are punctuation marks or special tokens with little lexical meaning, suggesting that attention-sink formation is largely independent of token semantics.

\section{Implementation Details of Observations} \label{apdx: obsdetails}

\subsection{Implementation of Heatmap of Induced Consistency.}\label{apdx: obsdetails1}
We conduct this observation on LLaDA-8B-Instruct using 100 randomly sampled instances from GSM8K and 100 from HumanEval, all decoded under vanilla (top-1) sampling with generation length $L=256$, block length $32$, and temperature $0$. 

For each request, we first run decoding to completion using vanilla sampling and store the resulting fully unmasked sequence as the reference output. We then replay the same trajectory and, at every denoising step, log per-edge statistics as follows. We form a sparse directed dependency graph as Sec.~\ref {meth: dgc}. The anchor is restricted to positions and prompt and active blocks, and the induced position is restricted to currently masked positions. For each retained edge, we record the anchor confidence, the induced confidence, and a binary consistency label indicating whether the induced position's current argmax prediction already matches the token at the same position in the reference output.

Records are binned in the (anchor confidence, induced confidence) plane with width $0.05$. Since the same anchor--induced pair recurs across consecutive steps while the induced position remains masked, we deduplicate within each bin by retaining only the earliest observation, which provides the most conservative test of consistency at that confidence level. The reported consistency ratio per bin is the fraction of retained records with a positive consistency label, and bins with fewer than $50$ records are rendered in gray.

\subsection{Implementation of Influence Between Masked Positions.}\label{apdx: obsdetails2}
We conduct this observation on LLaDA-8B-Instruct using 100 randomly sampled instances from GSM8K and 100 from HumanEval, decoded under vanilla (top-1) sampling with generation length $L=256$, block length $32$, and temperature $0$. The goal is to assess whether directed edges in our attention-derived dependency graph capture genuine influence: does unmasking the source of an edge induce a larger distributional shift at the target than unmasking an unrelated position?

At nine evenly spaced progress points from 10\% to 90\% of generation completion per request, we form a sparse directed dependency graph as in Sec.~\ref{meth: dgc} and construct a candidate pool $R$ of currently masked positions. Unordered pairs from $R$ are partitioned into \emph{coupled pairs} (those connected by at least one directed edge in the dependency graph) and \emph{non-coupled pairs} (those with no edge between them); coupled pairs are capped at 50 per step, and non-coupled pairs are subsampled to match.

For each pair $(p_a, p_b)$, we choose which position $p_i$ to unmask first: for coupled pairs, $p_i$ is set to the edge source (uniformly random if both directions are present); for non-coupled pairs, $p_i$ is chosen uniformly at random. Let $p_j$ denote the remaining position. We compare the model's existing prediction at $p_j$ ($P_{\text{joint}}$) with the prediction obtained after committing $p_i$ to its argmax and performing a fresh forward pass ($P_{\text{seq}}$), recording $\mathrm{KL}(P_{\text{joint}} \,\Vert\, P_{\text{seq}})$ as the per-pair coupling measure. All sequential forwards operate on cloned tensors and do not affect the underlying decoding trajectory. Reported statistics are the median KL and its interquartile range, computed separately for coupled and non-coupled pairs.

\subsection{Additional Validation of Masked-Position Influence}\label{apdx:additional_obs}

\begin{figure*}[t!]
    \centering
    \includegraphics[width=\textwidth]{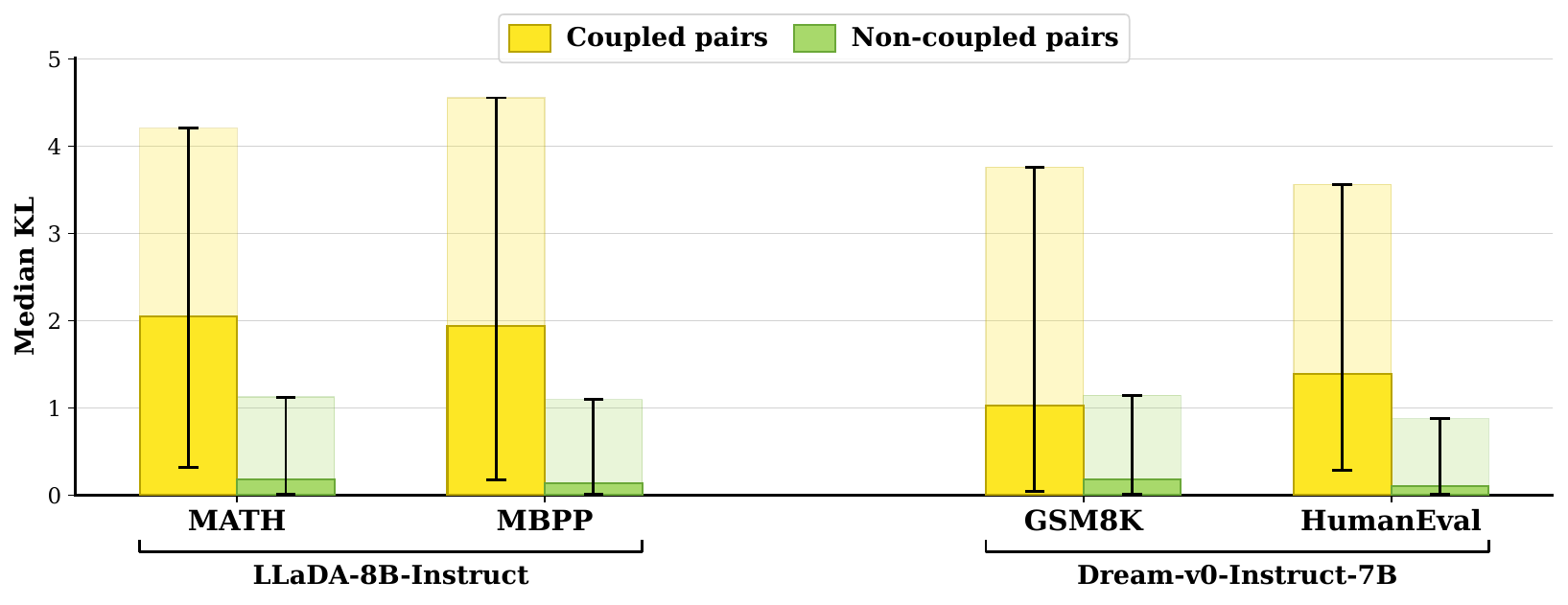}
    \caption{\textbf{Additional Validation of Influence Between Masked Positions.} Coupled pairs consistently exhibit larger KL divergence than non-coupled pairs, further validating the attention-derived dependency relations.}
    \label{fig:additional_obs}
\end{figure*}

To further strengthen our validation, we have added results on additional datasets and on the Dream model, which show the further generality. Fig.~\ref{fig:additional_obs} shows KL statistics consistent with those in Fig.~\ref{fig:obs} (right), showing that coupled pairs exhibit substantially stronger mutual influence than non-coupled pairs. These additional results provide broader evidence for the reliability of the dependency proxy.

Rigorously establishing the mechanism underlying attention is beyond the scope of this work and would require substantial additional work. We therefore phrase our claims conservatively, ground in prior studies of attention mechanisms, and provide sufficient evidence for the specific properties required by our method. All settings consistently support our central claim that the proposed dependency proxy identifies token pairs with strong mutual influence. 

\section{Experiment Details} \label{apdx: expdetails}

\subsection{Hyperparameter Settings}\label{apdx: hyper}
We conduct our experiments on several variants of two models: LLaDA-8B-Instruct, LLaDA-1.5, Dream-v0-Base-7B, Dream-v0-Instruct-7B. Benchmarks include diverse datasets: GSM8K (5-shot), MATH (4-shot), HumanEval (0-shot), and MBPP (3-shot), covering a range of reasoning and code generation tasks. Across all settings, we fix the block size to 32 and generation length to 256 tokens except KLASS, which uses its best-performing block length.

\begin{figure}[t!]
    \centering
    \includegraphics[width=\linewidth]{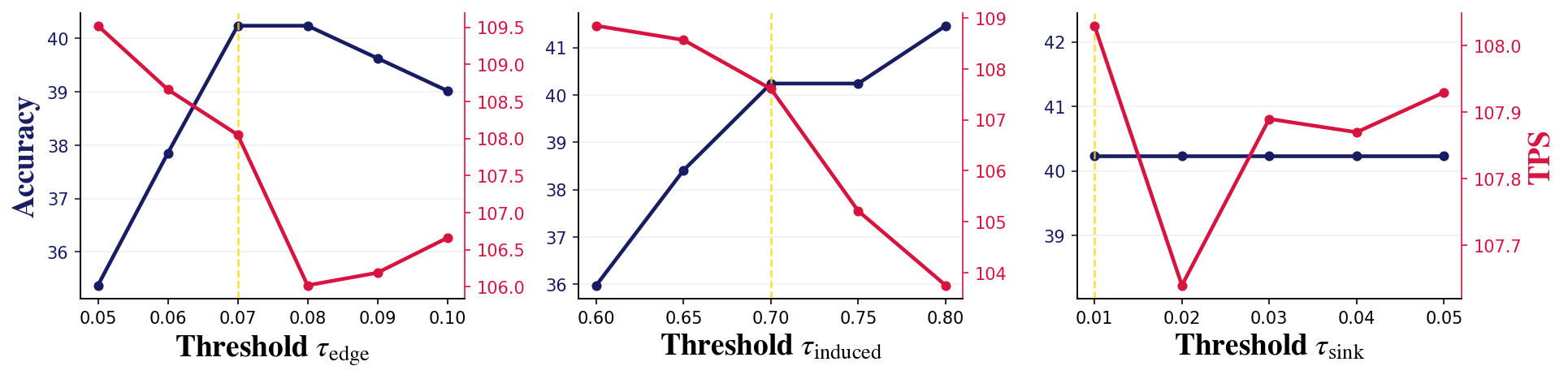}
    \caption{\textbf{We vary the threshold $\tau_{edge}$, $\tau_{induced}$, $\tau_{sink}$ and report accuracy (blue, left y-axis) and throughput TPS (red, right y-axis) on LLaDA-8B-Instruct}. The dashed lines (yellow) mark the final settings $\tau_{edge}$ = 0.07, $\tau_{induced}$ = 0.70, $\tau_{sink}$ = 0.01.}
    \label{fig:llada-8b-instruct}
\end{figure}

\begin{figure}[t!]
    \centering
    \includegraphics[width=\linewidth]{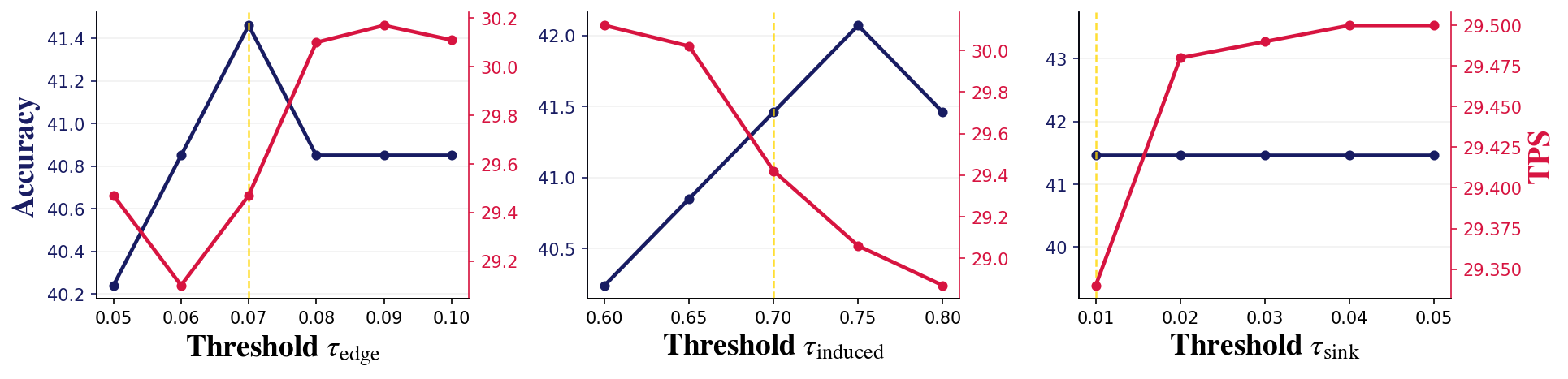}
    \caption{\textbf{We vary the threshold $\tau_{edge}$, $\tau_{induced}$, $\tau_{sink}$ and report accuracy (blue, left y-axis) and throughput TPS (red, right y-axis) on LLaDA-1.5}. The dashed lines (yellow) mark the final settings $\tau_{edge}$ = 0.07, $\tau_{induced}$ = 0.70, $\tau_{sink}$ = 0.01.}
    \label{fig:llada-1.5}
\end{figure}

\begin{figure}[t!]
    \centering
    \includegraphics[width=\linewidth]{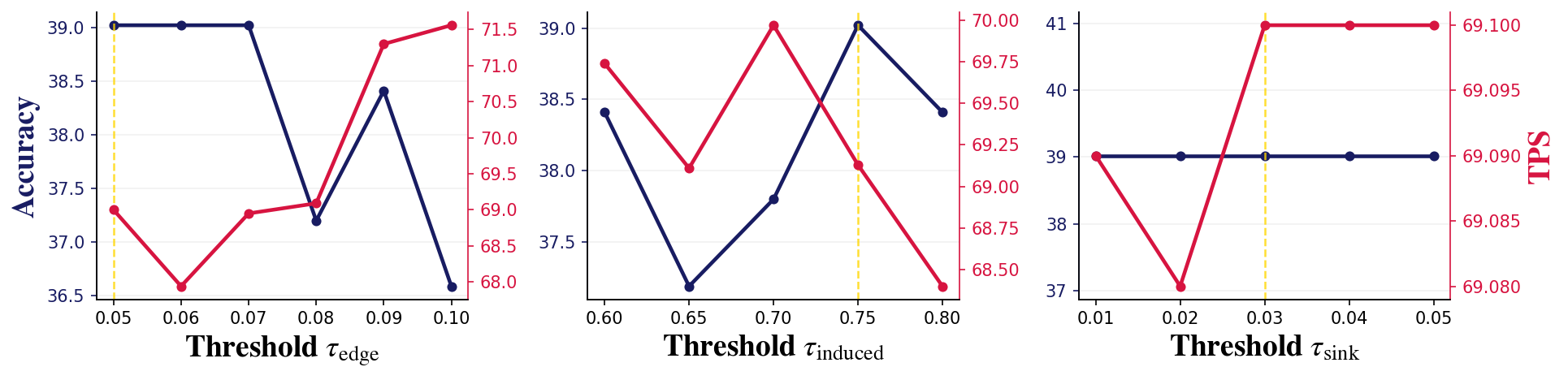}
    \caption{\textbf{We vary the threshold $\tau_{edge}$, $\tau_{induced}$, $\tau_{sink}$ and report accuracy (blue, left y-axis) and throughput TPS (red, right y-axis) on Dream-v0-Base-7B}. The dashed lines (yellow) mark the final settings $\tau_{edge}$ = 0.05, $\tau_{induced}$ = 0.75, $\tau_{sink}$ = 0.03.}
    \label{fig:dream-base}
\end{figure}

\begin{figure}[t!]
    \centering
    \includegraphics[width=\linewidth]{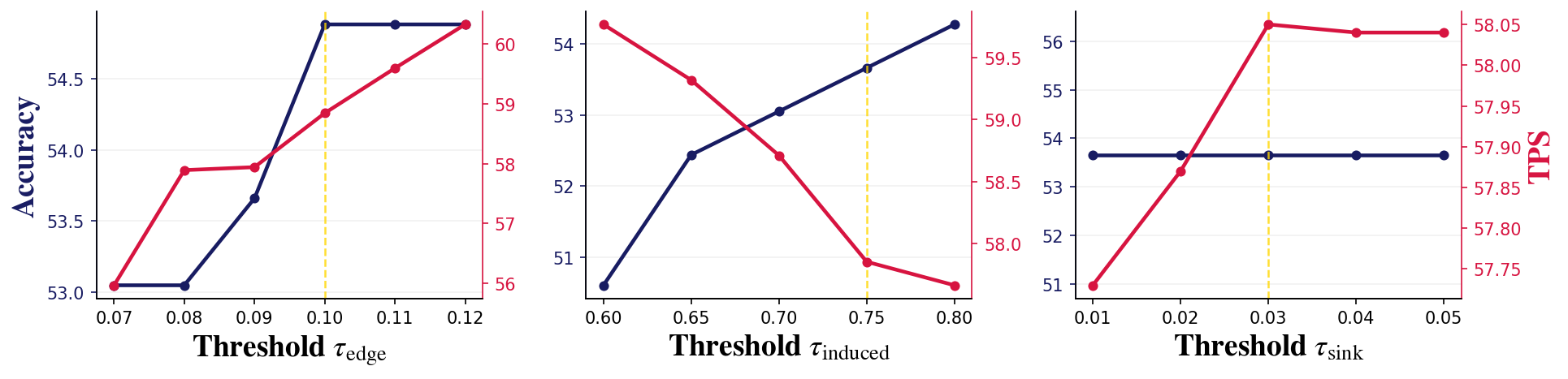}
    \caption{\textbf{We vary the threshold $\tau_{edge}$, $\tau_{induced}$, $\tau_{sink}$ and report accuracy (blue, left y-axis) and throughput TPS (red, right y-axis) on Dream-v0-Instruct-7B}. The dashed lines (yellow) mark the final settings $\tau_{edge}$ = 0.10, $\tau_{induced}$ = 0.75, $\tau_{sink}$ = 0.03.}
    \label{fig:dream-instruct}
\end{figure}

All baselines are evaluated under their default hyperparameter settings. To identify the optimal hyperparameter configuration for our method \dawn{} on different models, including $\tau_{edge}$, $\tau_{induced}$ and $\tau_{sink}$, we conduct a grid search on the HumanEval dataset. Specifically, we evaluate a range of candidate values for each parameter and plot the corresponding Accuracy–TPS curves, visualized in Fig.~\ref{fig:llada-8b-instruct}, \ref{fig:llada-1.5}, \ref{fig:dream-base}, \ref{fig:dream-instruct}.

In most cases, we observe a clear trade-off between throughput and accuracy: higher throughput is generally achieved at the cost of reduced accuracy. To balance this trade-off, we select hyperparameter values that lie near the Pareto frontier, favoring configurations that preserve high accuracy while providing meaningful efficiency gains. For example, on the LLaDA-8B-Instruct model, we select $\tau_{edge} = 0.07$, which achieves the highest accuracy among the tested values while offering higher throughput compared to $\tau_{edge} = 0.08$. This criterion ensures that the chosen configuration does not sacrifice accuracy for marginal speed improvements.

For \textit{the lower confidence threshold $\tau_{low}$} hyperparameter, we perform a grid search over both models and benchmarks, resulting in 16 experimental settings in total. Within each setting, we further tune the method-specific hyperparameters to obtain the best-performing configuration. The final selected results are reported in Tab.~\ref{tab:4x4}.


However, fine-grained tuning of these thresholds introduces additional overhead. \dawn{} introduces several thresholds, including $\tau_{\text{edge}}$, $\tau_{\text{induced}}$, $\tau_{\text{sink}}$, and $\tau_{\text{low}}$, whose optimal values were lightly tuned in our main experiments to demonstrate the upper-bound performance of the method. Among them, $\tau_{\text{low}}$ is by far the most influential: variations in this threshold lead to substantial changes in the quality--speed trade-off, while the remaining thresholds exhibit much smaller sensitivity. In practice, users may focus tuning effort on $\tau_{\text{low}}$ alone and adopt fixed defaults for the others. Specifically, we recommend $\tau_{\text{induced}}=0.7$, $\tau_{\text{sink}}=0.01$, and $\tau_{\text{edge}}=0.07$ for the LLaDA family of models, and $\tau_{\text{induced}}=0.75$, $\tau_{\text{sink}}=0.03$, $\tau_{\text{edge}}=0.07$ for the Dream family. These defaults yield results comparable to fully tuned configurations across our benchmarks, suggesting that the practical tuning burden is light despite the apparent number of hyperparameters.

\begin{table*}[t!]
  \centering
  \caption{\textbf{Final hyperparameter settings for \dawn{}, where rows denote models and columns denote hyperparameter types}. The $\tau_{induced}$, $\tau_{sink}$, $\tau_{edge}$ are shared across benchmarks for each model, while $\tau_{low}$ adopts benchmark-specific configurations.}
  \begin{tabular}{*{8}{c}}
    \toprule
    & \multirow{2}{*}{\textbf{$\tau_{induced}$}} & \multirow{2}{*}{\textbf{$\tau_{sink}$}} & \multirow{2}{*}{\textbf{$\tau_{edge}$}} & \multicolumn{4}{c}{$\tau_{low}$} \\
    \cmidrule{5-8}
    Model & & & & GSM8K & MATH & HumanEval & MBPP \\
    \midrule
    LLaDA-8B-Instruct     & 0.70 & 0.01 & 0.07 & 0.75 & 0.75 & 0.8 & 0.7\\
    LLaDA-1.5             & 0.70 & 0.01 & 0.07 & 0.75 & 0.75 & 0.8 & 0.75\\
    Dream-v0-Base-7B       & 0.75 & 0.03 & 0.05 & 0.75 & 0.8 & 0.8 & 0.8\\
    Dream-v0-Instruct-7B   & 0.75 & 0.03 & 0.10 & 0.8 & 0.8 & 0.8 & 0.8\\
    \bottomrule
  \end{tabular}
  \label{tab:4x4}
\end{table*}

\subsection{Additional Evaluation on IFEval} \label{apdx:ifeval}

\begin{table*}[t!]
  \centering
  \small
  \caption{\textbf{Performance comparison between \dawn{} and baselines on IFEval}. We report Accuracy and TPS to assess generation quality and efficiency.}
  \label{tab:ifeval}
  \resizebox{\textwidth}{!}{
    \begin{tabular}{c*{2}{cccc}}
      \toprule
      Metric
        & Vanilla & Confidence & LocalLeap & \dawn{}
        & Vanilla & Confidence & LocalLeap & \dawn{}\\
      \midrule
        & \multicolumn{4}{c}{LLaDA-8B-Instruct, $\tau_{low}=0.75$}
        & \multicolumn{4}{c}{LLaDA-1.5, $\tau_{low}=0.8$}\\
      \midrule
      Acc.
        & 56.75 & 57.12 & 57.30 & 56.19
        & 54.53 & 54.71 & 54.52 & 55.27\\
      TPS
        & 21.52 & 40.64 & 44.10 & \textbf{45.26}
        & 21.68 & 40.17 & 43.23 & \textbf{43.33}\\
      \midrule
        & \multicolumn{4}{c}{Dream-v0-Base-7B, $\tau_{low}=0.75$}
        & \multicolumn{4}{c}{Dream-v0-Instruct-7B, $\tau_{low}=0.8$}\\
      \midrule
      Acc.
        & 22.55 & 21.81 & 21.63 & 21.81
        & 27.54 & 28.47 & 28.10 & 27.54\\
      TPS
        & 37.28 & 53.28 & 55.28 & \textbf{59.37}
        & 23.74 & 43.94 & 45.90 & \textbf{47.42}\\
      \bottomrule
    \end{tabular}
  }
\end{table*}

In fact, Reasoning and coding are among the primary task categories for dLLMs. To further broaden the evaluation scope, we additionally report results on IFEval (0-shot) \citep{ifeval}, a general instruction-following, open-ended generation benchmark. KLASS is not included as no specific hyperparameter setting for this task. The setting of \dawn{} follows the main experiments. Tab.~\ref{tab:ifeval} demonstrates that \dawn{} remains effective on instruction-following and open-ended generation benchmarks, improving inference efficiency while preserving output quality.

\begin{table*}[t!]
  \centering
  \small
  \caption{\textbf{Error-bar analysis of \dawn{} and representative baselines across 2 datasets and 4 models}. Results are reported as mean $\pm$ standard deviation over three runs.}
  \label{tab:statistical_analysis}
  \resizebox{\textwidth}{!}{
    \begin{tabular}{c*{4}{cc}}
      \toprule
        & \multicolumn{2}{c}{GSM8K}
        & \multicolumn{2}{c}{HumanEval}
        & \multicolumn{2}{c}{GSM8K}
        & \multicolumn{2}{c}{HumanEval}\\
      \cmidrule(lr){2-3} \cmidrule(lr){4-5}
      \cmidrule(lr){6-7} \cmidrule(lr){8-9}
      Method
        & Acc. & TPS
        & Acc. & TPS
        & Acc. & TPS
        & Acc. & TPS\\
      \midrule
        & \multicolumn{4}{c}{LLaDA-8B-Instruct}
        & \multicolumn{4}{c}{LLaDA-1.5}\\
      \midrule
      Vanilla
        & 77.94 & 10.40 $\pm$ 0.08
        & 40.24 & 26.73 $\pm$ 0.27
        & 81.12 & 9.73 $\pm$ 0.07
        & 43.90 & 9.29 $\pm$ 0.03\\
      LocalLeap
        & 77.33 & 44.64 $\pm$ 0.43
        & 39.63 & 109.9 $\pm$ 0.63
        & 80.14 & 42.88 $\pm$ 0.24
        & 42.07 & 30.44 $\pm$ 0.20\\
      \dawn{}
        & 77.94 & 45.12 $\pm$ 0.37
        & 40.24 & 109.2 $\pm$ 0.70
        & 80.82 & 43.46 $\pm$ 0.19
        & 42.07 & 29.72 $\pm$ 0.17\\
      \midrule
        & \multicolumn{4}{c}{Dream-v0-Base-7B}
        & \multicolumn{4}{c}{Dream-v0-Instruct-7B}\\
      \midrule
      Vanilla
        & 76.42 & 14.41 $\pm$ 0.03
        & 38.41 & 33.55 $\pm$ 0.18
        & 76.35 & 7.320 $\pm$ 0.02
        & 53.66 & 23.00 $\pm$ 0.30\\
      LocalLeap
        & 72.55 & 25.68 $\pm$ 0.24
        & 36.59 & 69.11 $\pm$ 0.72
        & 73.24 & 33.05 $\pm$ 0.10
        & 54.27 & 59.71 $\pm$ 0.38\\
      \dawn{}
        & 73.54 & 25.68 $\pm$ 0.19
        & 39.63 & 69.68 $\pm$ 0.66
        & 73.16 & 33.07 $\pm$ 0.08
        & 54.88 & 60.98 $\pm$ 0.65\\
      \bottomrule
    \end{tabular}
  }
\end{table*}

\subsection{Impact of Graph Sparsity}
When the dependency graph is sparse, AGD induces fewer masked positions due to the reduced number of dependency relations, while CBS becomes less constrained by graph edges and can therefore select more tokens for decoding. Conversely, when the graph is dense, AGD induces more masked positions, whereas CBS becomes more conservative because of the stronger dependency constraints. The parameter $\tau_{\text{edge}}$ controls the sparsity of the dependency graph at a global level: larger values produce sparser graphs. The ablation study of $\tau_{\text{edge}}$ in these figures shows that performance varies smoothly across different graph sparsities, without any abrupt degradation, demonstrating that \dawn{} is not overly sensitive to the choice of graph sparsity.

\section{Statistical Analysis} \label{statistical}

Most dLLM methods evaluated in our study use greedy decoding and are therefore deterministic with respect to the random seed. KLASS is the only stochastic baseline; however, given its substantial performance gap, we focus our robustness analysis on the strongest and most directly comparable baselines. To further verify the reliability of our results, we repeated representative experiments three times and report the error bar for vanilla decoding, LocalLeap, and \dawn{} across four models and two datasets.

Tab.~\ref{tab:statistical_analysis} confirms the reliability of the results reported in the main experiments. The mean values lead to the same conclusions as the main results, while the small standard deviations indicate that performance remains stable across repeated runs. Overall, \dawn{} consistently improves inference efficiency while maintaining comparable accuracy.



\end{document}